\documentclass[sn-mathphys]{sn-jnl}% Math and Physical Sciences Reference Style
\usepackage{graphicx}%
\usepackage{multirow}%
\usepackage{amsmath,amssymb,amsfonts}%
\usepackage{amsthm}%
\usepackage{mathrsfs}%
\usepackage[title]{appendix}%
\usepackage{xcolor}%
\usepackage{textcomp}%
\usepackage{manyfoot}%
\usepackage{booktabs}%
\usepackage{algorithm}%
\usepackage{algorithmicx}%
\usepackage{algpseudocode}%
\usepackage{listings}%

\usepackage{amsmath}
\usepackage{amssymb}
\usepackage{multirow}
\usepackage[title]{appendix}
\usepackage{etoolbox}
\usepackage[T1]{fontenc}
\usepackage{cprotect}
\usepackage{color}
\usepackage{enumitem}
\usepackage{bbm}
\usepackage{hyperref}
\usepackage{amsfonts}

\newcommand{\numpapers}{72 }
\newtheorem{definition}{Definition}

\newlist{questions}{enumerate}{2}
\setlist[questions,1]{label=\textbf{RQ\arabic*.},ref=RQ\arabic*}
\setlist[questions,2]{label=(\alph*),ref=\thequestionsi(\alph*)}

\begin{document}

\title[Modeling Risk in Reinforcement Learning: A Literature Mapping]{Modeling Risk in Reinforcement Learning: A Literature Mapping}


\author*{\fnm{Leonardo} \sur{Villalobos-Arias}}\email{lvillal@ncsu.edu}

\author{\fnm{Derek} \sur{Martin}}\email{dmartin7@ncsu.edu}

\author{\fnm{Abhijeet} \sur{Krishnan}}\email{akrish13@ncsu.edu}

\author{\fnm{Madeleine} \sur{Gagné}}\email{mmgagne@ncsu.edu}

\author{\fnm{Colin M.} \sur{Potts}}\email{cmpotts@ncsu.edu}

\author{\fnm{Arnav} \sur{Jhala}}\email{ahjhala@ncsu.edu}

\affil{\orgdiv{Department of Computer Science}, \orgname{North Carolina State University}, \orgaddress{\city{Raleigh}, \state{NC}, \country{USA}}}


\abstract{
% Background
Safe reinforcement learning deals with mitigating or avoiding unsafe situations by reinforcement learning (RL) agents.
% Problem
Safe RL approaches are based on specific risk representations for particular problems or domains. In order to analyze agent behaviors, compare safe RL approaches, and effectively transfer techniques between application domains, it is necessary to understand the types of risk specific to safe RL problems. 
% Objective
We performed a systematic literature mapping with the objective to characterize risk in safe RL. Based on the obtained results, we present definitions, characteristics, and types of risk that hold on multiple application domains.
% Method
Our literature mapping covers literature from the last 5 years~(2017-2022), from a variety of knowledge areas (AI, finance, engineering, medicine) where RL approaches emphasize risk representation and management.
% Results
Our mapping covers \numpapers papers filtered systematically from over thousands of papers on the topic. Our proposed notion of risk covers a variety of representations, disciplinary differences, common training exercises, and types of techniques. %, and action/policy preferences. \comment{The last one is true, but we did not go so much in depth.}
% Conclusion
We encourage researchers to include explicit and detailed accounts of risk in future safe RL research reports, using this mapping as a starting point. With this information, researchers and practitioners could draw stronger conclusions on the effectiveness of techniques on different problems.
}




\maketitle

\section{Introduction}
\label{sec:intr}

The reinforcement learning~(RL) framework consists of learning an optimal policy that maximizes the obtained reward by interacting with an environment~\citep{sutton2018reinforcement}. While effective, this puts the agent at risk in scenarios that feature dangerous or unsafe conditions, like many real-life situations. Robotic control tasks performed by an agent with little experience can damage or destroy its equipment. Engaging in risky or sub-optimal behavior in scenarios like portfolio management results in loss of assets or profit~\citep{amodei2016concrete}. To counteract this, safe reinforcement learning extends traditional RL to explicitly model the safety of the agent and its actions in the environment~\citep{garcia2015comprehensive}. Safe RL has been researched in domains where such decisions have a significant impact on human interaction such as autonomous driving~\citep{kiran2021deep}, robotics~\citep{brunke2022safe}, healthcare~\citep{yu2021reinforcement}, and economics~\citep{charpentier2021reinforcement}.

% Paragraph about how Safe RL is bad because its limited
Risk has been represented in RL as loss, variance, or stochasticity of the RL agent's reward~\citep{garcia2015comprehensive}. We seek recognition that there may be a more nuanced treatment of risk in terms of type, target of harmful actions, source of harmful actions, severity of real-life repercussions, frequency, and timing. While effective for solving optimization problems, the burden of figuring out the correct reward values and policy optimization criteria is often left to the domain experts. Moreover, techniques in RL are designed with different types of problems in mind. However, there is the possibility that techniques that are effective in one application domain can be useful in others, or made effective with some small changes. This could be the case if the problems in question exhibit similar types of risk. However, to the best of our knowledge, the literature on safe RL has not explicitly defined and addressed risk as a broad phenomenon, nor the possibility that risk is different from problem to problem.

% We contribute to the discussion of Safe RL through this survey track paper by providing a systematic mapping of existing RL approaches across domains.%To the best of our knowledge, there is no methodology or guidelines for designing reward functions for RL agents.

To address this problem, we perform a systematic literature mapping study in the area of safe reinforcement learning with the objective of identifying and characterizing risk. Our contribution is a definition of risk that fits with the current literature, with a way to identify, characterize and label different factors that contribute to the consideration of risk that a safe RL agent is tasked with. We present attributes and values that risk factors can have, regardless of the application domain. We moreover study the types of problems and representations of risk that are common in safe RL.

The rest of this paper is structured as follows. Section~\ref{sec:back} introduces the concepts of reinforcement learning and safe reinforcement learning necessary to understand our proposed definition of risk. Section~\ref{sec:meth} shows the methodology followed to perform the systematic mapping study presented in this article. Section~\ref{sec:defin} presents our definition of risk. Section~\ref{sec:chara} expands on this definition by describing a set of attributes that characterize risk. Section~\ref{sec:resu} showcases the results of our literature mapping and answers the question of how risk is characterized in safe RL. Section~\ref{sec:disc} highlights some of the most important findings of our mapping, including potential areas of future work. Lastly, section~\ref{sec:conc} echoes the main contributions of this paper with some closing thoughts.

\section{Safe Reinforcement Learning}
%% Background, basically
\label{sec:back}

Reinforcement learning algorithms work on the framework that an agent interacts with the environment by observing its state, selecting an action, and receiving a reward. In RL, the world is modeled using a Markov Decision Process~(MDP), defined by a tuple $M = \langle S,A,T,R \rangle$~\citep{sutton2018reinforcement}, where
\begin{itemize}
    \item $S$ is the state space,
    \item A is the action space,
    \item $T\colon S \times A \times S \to [0,1]$ is the transition function, and
    \item $R\colon S \times A \times S \to \mathbb{R}$ is the reward function
\end{itemize}

The aim of RL algorithms is to derive a \textit{policy} i.e., a mapping of states to actions $\pi\colon S \to A$ that optimizes for some criterion. Traditional RL optimization criteria are functions of the reward function R, such as expected cumulative future discounted reward, maximum worst-case return, the variance in return of the reward, cumulative value at risk~(CVaR), or combinations of these.

Safe Reinforcement Learning incorporates the safety of the agent to the MDP where the algorithm calculates the risk posed to the agent for performing an action. This characterization of risk is traditionally domain-specific. Previous works have highlighted that risk can be related to the stochasticity of the environment~\citep{yang2022safety}, error states~\citep{kazantzidis2022train}, or cost functions or constraints~\citep{yang2021nlp}. These are the representations of risk elements in the environment, which are then represented in the heuristic or rewards. Instances of these elements include news articles that may affect operation~\citep{aboutorab2022reinforcement}, the collision of a robot in an environment~\citep{chen2021safe}, and the number of failed landings~\citep{turchetta2020safe}.

There appears no general definition of risk or the types of risk in the safe RL literature. This literature study thus proposes a unified, domain-agnostic treatment of risk, based on the way risk is characterized and treated in the papers identified for detailed mapping based on a systematic methodology. This general view of risk offers the potential to develop domain-independent strategies and evaluations for handling different types of risk.

\section{Methodology}
\label{sec:meth}

We present a systematic mapping study in the context of safe reinforcement learning, with the objective of characterizing the notion of risk in the reinforcement learning literature. We followed the guidelines for systematic literature mappings by~\cite{petersen2008systematic,petersen2015guidelines}. The following research questions guided the mapping process:

\begin{questions}[itemsep=3pt,itemindent=18pt]
    \item What are the characteristics of risk in reinforcement learning literature?\label{rq:r1}
    
    \item What are the reinforcement learning application domains where risk is explicitly modeled?\label{rq:r2}
    
    \item How is risk represented in reinforcement learning?\label{rq:r3}
\end{questions}

We first started by looking at recent literature studies on reinforcement learning~\citep{brunke2022safe,lockwood2022review,tambon2022certify,anand2021safe,haider2021domain,kiran2021deep,kim2020safe}. These literature surveys pointed us to the state-of-the-art on reinforcement learning in their specific sub-fields, as well as the methodology of literature surveys prevalent in the reinforcement learning field. These studies provide a good list of sources as a starting point due to their thorough treatment of related work within their specific are of focus. We move beyond these surveys to more concretely map the approaches across domains, techniques, and types of risks.

%While very thorough on their own line of research, it is not sufficient to answer the above questions. Particularly, we cannot get a full picture of how risk is represented in reinforcement learning, and how that is different depending on the application domain. Regardless, these studies functioned as a guideline to construct the survey protocol.

\begin{figure}
    \centering  
    \includegraphics[width=0.95\linewidth]{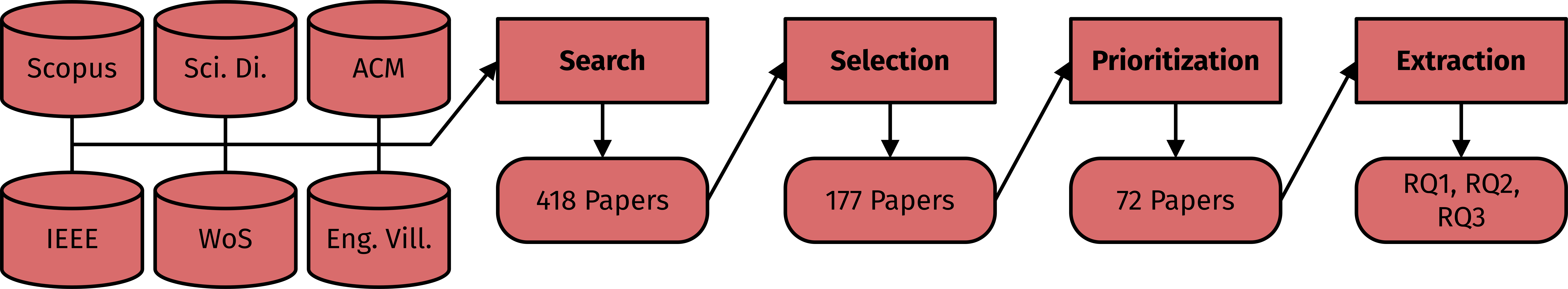}
    \caption{Summary of the systematic literature mapping process with steps and outcomes.}
    \label{fig:slm}
\end{figure}

Figure~\ref{fig:slm} summarizes the steps followed for this literature mapping.

\subsection{Search Strategy and Scoping}
We devised our search string from our research questions, plus a set of control articles in the area of risk-aware or safe reinforcement learning starting from foundational papers~\citep{heger1994consideration,borkar2001sensitivity,mihatsch2002risk,gaskett2003reinforcement,nilim2005robust,geibel2005risk,basu2008learning,tamar2012policy,geramifard2013intelligent,gehring2013smart,garcia2015comprehensive}. We worked and refined the search string until it could retrieve all articles in the control group, while simultaneously keeping a small number of broad keywords to provide a more comprehensive overview of the area. The final search string used to find relevant literature was the following:

\vspace{5pt}
{\center \texttt{`safe reinforcement learning' OR `safe RL' OR ((`reinforcement learning') AND (`risk sensitiv*' OR `risk manag*'))}}
\vspace{5pt}

To identify all relevant literature, we searched the following digital libraries: Scopus, ACM Digital Library, IEEE Xplore, Science Direct, Engineering Village (just Compendex), and Web of Science. We restricted our search to papers published in the range Jan 2017--Aug 2022 written in English. There were two reasons for this choice. First, there were fewer RL papers with practical applications that explicitly model safety in RL prior to this period. Second, other surveys provide better coverage of older papers. We also chose conference and journal articles in computer science or engineering or closely related areas. This was to ensure that we cover papers that formally use the term rather than with a more informal usage (this is apparent in Economics and Finance papers). The results were aggregated and duplicates were automatically removed based on the paper title, resulting in {\em 418} unique papers from the search result.

We applied inclusion and exclusion criteria to select only relevant research further based on a manual reading of the papers. To be included in this mapping, papers had to (I1)~consider safety, uncertainty, or some other notion of risk to an extent that it affected the problem statement or solution, (I2)~propose a novel RL approach or extend an existing one for a specific application, (I3) have at least one primary study (i.e., an experiment) to test the approach, and (I4) be published in an AI-related venue. We filtered out documents that were (E1)~not peer-reviewed publications (books, notes, slides), and (E2)~non-primary studies like literature reviews and discussion articles. The criteria were manually applied based on the title, keywords, and abstract. If the information in them was not sufficient, the full text of the article was used. Table~\ref{tab:papers} shows the \numpapers papers reviewed for this literature mapping. We use the ID numbers displayed on the table instead of citations on the numerous figures on this paper.

\begin{table}[t]
\caption{List of surveyed publications.}
\label{tab:papers}
\begin{tabular}{@{}rlrl@{}}
\toprule
\multicolumn{1}{l}{ID} & Reference                        & \multicolumn{1}{l}{ID} & Reference                         \\ \midrule
1                      & \cite*{zhang2022barrier}         & 37                     & \cite*{kanellopoulos2021temporal} \\
2                      & \cite*{mazouchi2021conflict}     & 38                     & \cite*{okawa2020automatic}        \\
3                      & \cite*{lim2022dynamic}           & 39                     & \cite*{keramati2020being}         \\
4                      & \cite*{kazantzidis2022train}     & 40                     & \cite*{stooke2020responsive}      \\
5                      & \cite*{garcia2022instance}       & 41                     & \cite*{yang2020safe1}              \\
6                      & \cite*{hansmeier2022integrating} & 42                     & \cite*{yang2020safe2}              \\
7                      & \cite*{den2022planning}          & 43                     & \cite*{wachi2020safe}             \\
8                      & \cite*{lenaers2022regular}       & 44                     & \cite*{jansen2020safe}            \\
9                      & \cite*{bisi2022risk}             & 45                     & \cite*{serrano2020safe}           \\
10                     & \cite*{zhang2022safe}            & 46                     & \cite*{turchetta2020safe}         \\
11                     & \cite*{liu2022safe}              & 47                     & \cite*{qian2020safe}              \\
12                     & \cite*{yang2022safety}           & 48                     & \cite*{gros2020safe}              \\
13                     & \cite*{cowen2022samba}           & 49                     & \cite*{andersen2020safer}         \\
14                     & \cite*{xuan2022sem}              & 50                     & \cite*{garcia2020teaching}        \\
15                     & \cite*{zanon2022stability}       & 51                     & \cite*{zaw2020verifying}          \\
16                     & \cite*{park2022uncertainty}      & 52                     & \cite*{yu2019convergent}          \\
17                     & \cite*{adel2021multi}            & 53                     & \cite*{cheng2019end}              \\
18                     & \cite*{simao2021alwayssafe}      & 54                     & \cite*{murugesan2019formal}       \\
19                     & \cite*{el2021abstraction}        & 55                     & \cite*{prakash2019improving}      \\
20                     & \cite*{perotto2021gambler}       & 56                     & \cite*{garcia2019probabilistic}   \\
21                     & \cite*{bedi2021intermittent}     & 57                     & \cite*{huang2019risk}             \\
22                     & \cite*{kong2021lane}             & 58                     & \cite*{reddy2019risk}             \\
23                     & \cite*{luo2021learning}          & 59                     & \cite*{simao2019safe}             \\
24                     & \cite*{cohen2021model}           & 60                     & \cite*{ge2019safe}                \\
25                     & \cite*{van2021no}                & 61                     & \cite*{ma2019state}               \\
26                     & \cite*{vamplew2021potential}     & 62                     & \cite*{chow2018lyapunov}          \\
27                     & \cite*{maeda2021reconnaissance}  & 63                     & \cite*{depeweg2018decomposition}  \\
28                     & \cite*{tlili2021risks}           & 64                     & \cite*{dilokthanakul2018deep}     \\
29                     & \cite*{wachi2021safe}            & 65                     & \cite*{pan2018efficient}          \\
30                     & \cite*{thomas2021safe}           & 66                     & \cite*{dabney2018implicit}        \\
31                     & \cite*{yang2021cps}              & 67                     & \cite*{fulton2018safe}            \\
32                     & \cite*{v2021safe}                & 68                     & \cite*{alshiekh2018safe}          \\
33                     & \cite*{yang2021nlp}              & 69                     & \cite*{verbist2018actor}          \\
34                     & \cite*{zhang2021safe}            & 70                     & \cite*{pathak2018verification}    \\
35                     & \cite*{marvi2021safe}            & 71                     & \cite*{lee2017constrained}        \\
36                     & \cite*{belzner2021synthesizing}  & 72                     & \cite*{saunders2017trial}         \\ \bottomrule
\end{tabular}
\end{table}

\subsection{Data extraction and synthesis}
Table~\ref{tab:ext} depicts the data extraction main categories used for the literature mapping. This breakdown of categories further facilitates assessments of risk and their cross-disciplinary classifications. For \ref{rq:r1}, we labeled the risks managed by each paper using the ontology presented in Section~\ref{sec:chara}. For \ref{rq:r2}, we identified the experiments and their associated MDPs, datasets, or descriptions. We then labeled each of these cases with one of five identified application domains: computer science, (non-CS) engineering, economics, medicine, and toy problems. Lastly, for \ref{rq:r3}, we identified the safe RL approach according to the comprehensive risk taxonomy presented in \cite{garcia2015comprehensive}. %The overall mapping of papers based on RQ2 experiments across domains and techniques is presented in Figure~\ref{fig:experiment-counts}.

\begin{table}[t]
    \caption{Data extraction categories.}
    \label{tab:ext}
    \centering
    \begin{tabular}{ll}
        \toprule
        RQ & Category \\
        \midrule
        1 & Risk types and characteristics \\
        2 & Experiment, domain \\
        3 & Technique, risk taxonomy classification \\
        \bottomrule
    \end{tabular}
    
\end{table}

The methodology, search, extraction process, and results of each research question are available in the following repository:~\url{http://tiny.cc/riskmapping}.

\section{Risk in Reinforcement Learning}
\label{sec:defin}

We start by recounting general definitions of RL terms that allow us to express risk characterization and assessment in a framework that can represent different task and application domains.
%We propose the following definitions for the generality of reinforcement learning. Because of the specifics that each application domain and experiment have when profiling risks, we provide a general framework for risk assessment and characterization. Because of this, our definition of risk will be based on the \textit{MDP}~(Section~\ref{sec:back}), and risk is externally defined by the user or designer.

Let $M = \langle S,A,T,R \rangle$ be a MDP, and transitions in $T: S \times A \times S$ be written as tuples $t = (s_{1}, a, s_{2})$. Let $\tau$ be the domain of all transitions in $M$ so that $t \in \tau \leftrightarrow T(s_{1}, a, s_{2}) > 0$. The origin state-action pair of a transition refers to $(s_{1}, a)$, and the resulting state refers to $s_{2}$.

\begin{definition}[Outcome]
    An outcome $O$ refers to a non-empty subset of $\tau$.
\end{definition}
Outcomes are groupings of transitions that have some external meaning. For example, a particular outcome can hold all of the transitions that can be visited by a policy $\pi$. That outcome can be thus defined by all transitions $(s_{1},a,s_{2})$ so that $\pi(s_{1}) = a$ and $T(s_{1},a,s_{2}) > 0$.

\begin{definition}[Utility]
    The utility $U$ of a transition is a user-defined function that assigns a numeric value to each transition. A transition with a higher utility is preferable to a lower utility transition.

    Similarly, the utility $V$ of a state is a user-defined function that assigns a numeric value to a state, with a higher utility being preferable.
    
    The utility of a transition is defined as the reward gained from the transition plus the utility of its resulting state: 
    
    $$
    U(t=(s_{1},a,s_{2}))=R(s_{1},a,s_{2})+V(s_{2}).
    $$
\end{definition}
The utility function is a way to account for the different representations and preferences of risk across different RL domains.

The utility of a state has been exhaustively studied in safe reinforcement learning, as different types of optimization functions. Below some examples of state utility are listed, but a more comprehensive list has been compiled by \cite{garcia2015comprehensive} and \cite{sutton2018reinforcement}.

\begin{itemize}
    \item Immediate reward: $V(s) = 0$
    \item Risk neutral: $V(s)=max_{\pi \in \Pi} E_{\pi}(\sum_{t=0}^{\infty} \gamma^{t} r_{t})$
    \item Risk-sensitive: $V(s)=max_{\pi \in \Pi} (E_{\pi}(\sum_{t=0}^{\infty} \gamma^{t} r_{t}) - \beta \omega)$, where $\beta$ is a risk preference parameter and $\omega$ is the representation of risk.
\end{itemize}

\begin{definition}[Domain Risk]
An outcome is a domain risk if, for every transition inside it, there exists another transition with the same origin state-action pair outside of the outcome with a higher utility

$$
\forall_{(s_{1},a,s_{2}) \in O} \exists_{s_{3} \in S}: (s_{1},a,s_{3}) \notin O \land U(s_{1},a,s_{2}) < U(s_{1},a,s_{3}).
$$

\end{definition}

Domain risks represent the intrinsic loss of expected utility that an RL agent may take when selecting an action. That is, an agent could select an action with the hopes that it could net a higher utility transition while having the possibility of ending on a lower utility transition. Domain risks require uncertainty about the outcome so that all origin state-action pairs can result in more than one state. This type of uncertainty is called aleatoric uncertainty.

Notably, an outcome is \textit{not} a domain risk when it includes a deterministic transition or a transition that is `optimal' (highest utility for the origin state-action pair). A domain risk is then a collection of externally defined `bad' transitions. It notably does not take into account the decision-making ability of the agent.

\begin{definition}[Policy Outcome]
    For an agent with a policy $\pi$, let $O_{\pi}$ be its policy outcome: a set of transitions that can be visited under that policy:

    $$
    (s_{1},\pi(s_1),s_{2}) \in O.
    $$
\end{definition}

\begin{definition}[Policy Risk]
    A policy outcome is a policy risk if it contains a transition for which there exists another transition with the same starting state outside of the outcome with a higher utility:

    $$
    \exists_{(s_{1},a_{1},s_{2}) \in O_{\pi}} \exists_{(s_{1},a_{2},s_{3}) \in \tau} : (s_{1},a_{2},s_{3}) \notin O_{\pi} \land U(s_{1},a_{1},s_{2}) < U(s_{1},a_{2},s_{3}).
    $$
\end{definition}

Policy risk represents the potential trade-off when an agent chooses one action over another when building its policy. Policy risk happens when there is another action that may result in higher utility than the one chosen. Notably, policy risks do not include cases where one action may result in different utilities; given this is a domain risk.

Policy risk is an internal definition, as a policy will or will not exhibit risk when selecting actions. The definition also includes policies that might not be optimal from a utility standpoint.

\begin{definition}[Confidence]
    Let $M'=\langle S',A',T',R' \rangle$ be the agent’s approximation of the original MDP. The confidence function $K(s,a) \rightarrow [0,1)$ represents a self-evaluation of the agent on how well $T'$ and $R'$ approximate the original $T$ and $R$. $K(s,a)=0$ when the agent has no knowledge about a state-action pair, and it approaches $1$ as the agent is more confident.
\end{definition}

An RL agent learns the qualities and distributions of the MDP through exploration. The agent's estimate of the MDP $M'$ will start out as a very loose model but will start resembling the optimal with sufficient experience. For most practical real-world domains, and MDP does not fully model the real-world. In such cases, there is uncertainty due to external factors that are outside of the agent's observations. This introduces a type of uncertainty that we label as \textit{epistemic} uncertainty.

%However, the agent will never have access to $M$, and thus will always have uncertainty with how well $M'$ approximates $M$. This type of uncertainty is known as epistemic uncertainty.

\begin{definition}[Exploration Risk]
    Let $k \in [0,1]$ be a confidence threshold, so that the agent considers $K(s,a) \geq k$ as a sufficiently explored state-action pair, and otherwise it is considered unexplored.
    
    Let $u \in \mathbb{R}$ be a utility threshold so that the agent considers $R(t) < u$ as an undesirable transition.

    An outcome is an Exploration Risk if each of its transitions is both unexplored and undesirable

    $$
        \forall_{(s_{1},a,s_{2}) \in O} : K(s_{1},a) < k \land U(s_{1},a,s_{2}) < u.
    $$
    
\end{definition}

Exploration risks represent undesirable transitions that an agent would like to avoid during exploration of the MDP, and will potentially consider during policy selection. A risk-aware agent will consider possible exploration risks when deciding to select an action.

Exploration risks are the only type of risk that changes, and they can be eliminated given enough agent knowledge. When enough knowledge and confidence are acquired about a state-action pair, the agent could decide whether to select (potentially becoming a policy risk) or avoid (stops being a risk).

Similar to domain risk, exploration risks are a collection of externally defined transitions with some undesirable connotation. The main difference is that exploration risks do not require aleatoric uncertainty, and are instead characterized by having lower utility values.

For instance, an agent falling into a hole can be considered an exploration risk as it can be an unfavorable result. In a deterministic environment, it would not constitute a domain risk, as the agent can completely avoid falling into the hole with enough experience.

Domain and policy risks can also be converted to exploration risks. Assuming the agent has zero knowledge of any state-action pairs, if we choose $k = \max\limits_{t \in O} U(t) + \epsilon$. In other words, all domain and policy risk are exploration risks.

\begin{definition}[Risk Factor]
    A risk factor is an externally defined outcome that is either a domain or exploration risk.
\end{definition}

Risk factors are the dangerous or undesirable features or mechanics of the environment that the agent should avoid or mitigate. Risk factors represent a group of transitions that have some relation, such as having the same cause, affected entity, timing, frequency, or worst-case scenario.

We exclude policy risks from the definition of risk factor, as 1) these have to do with the final selection of the optimal policy, 2) are only a consideration in the final step of reinforcement learning, and 3) are present in any non-trivial RL problem. As such, they do not constitute an interesting challenge from the perspective of profiling and characterizing risk.

The following are some examples of risk factors in the Safe RL literature:
\begin{itemize}
    \item An agent falls into a hole
    \item Potential loss of profit from investing in a stock
    \item The agent randomly slips and falls
    \item The agent randomly walks in an unintended direction
    \item A self-driving car intentionally crashes
    \item A self-driving car crashes due to imprecise controls
\end{itemize}

Because risk factors are externally defined, there is some subjectivity on what constitutes a risk factor for a particular domain and application. As such, we present some guidelines on how can researchers and practitioners of safe RL identify and characterize risks~(Section~\ref{sec:disc}). In the following section, we define the characteristics of risk factors. In addition, we highlight common types of risk factors in the Safe RL literature when presenting the results of the mapping study.

\section{Risk Factors}
\label{sec:chara}

This work characterizes risk factors under a set of attributes, as the nature of risk is different from one domain to another. Decision-making agents often need to assess trade-offs with respect to the type, likelihood, and severity of risk.

This work serves as an update and synthesis of previously published surveys on different aspects of safe reinforcement learning, such as one by~\cite{garcia2015comprehensive}, the optimization techniques survey by~\cite{kim2020safe}, the survey by~\cite{brunke2022safe} on safe learning in robotics, and the literature review on certification by~\cite{tambon2022certify}. Previous research allowed us to study risk more broadly. The proposed risk factor characteristics provide a vocabulary to map the articles we found in our literature mapping in terms of types of risk, representations, and application domain characteristics.

\begin{figure}[ht]
    \centering
    \includegraphics[width=0.9\linewidth]{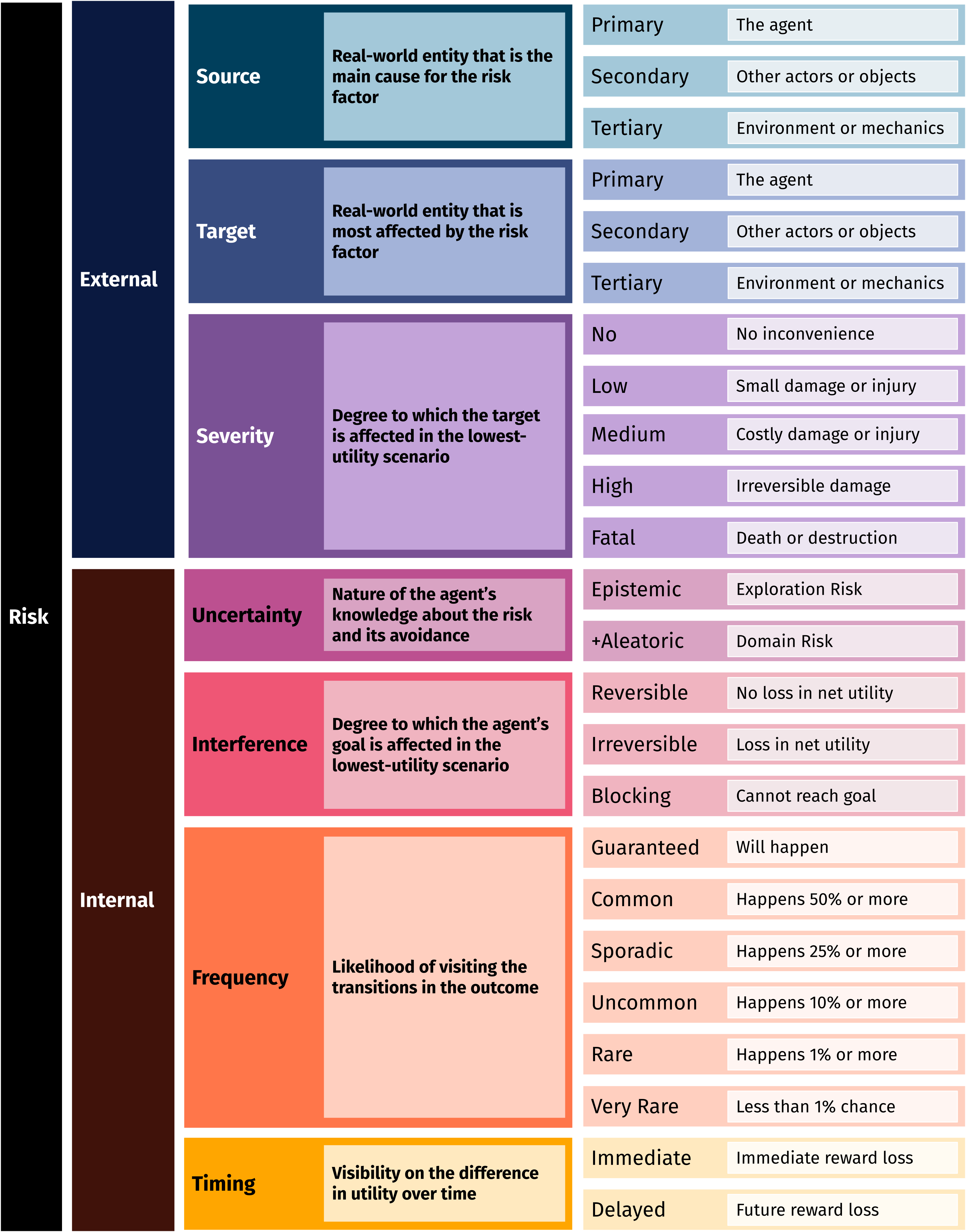}
    \caption{Summary of risk factor characteristics and values.}
    \label{fig:risk-chara}
\end{figure}

We define the following attributes: \textit{source}, \textit{target}, \textit{severity}, \textit{uncertainty}, \textit{interference}, and \textit{frequency}, and \textit{timing}. Risk factors have one value per attribute, from a set of possible values. Figure~\ref{fig:risk-chara} shows a summary of the characteristics of risk factors and their possible values.

Because of the nature of risk factors as externally defined outcomes, there are characteristics of risk factors that are not captured in the MDP specification. Cause and effect relationships and allegories to the real world are lost when formulating a problem as a reinforcement learning task. Thus, a portion of the risk factor characteristics are \textit{external}, meaning they are not derivable from the outcome itself, but rather from the problem being modeled. On the other hand \textit{internal} characteristics are derivable from the outcome (the set of transitions). The source, target, and severity are external characteristics of a risk factor, and the uncertainty, interference, frequency, and timing are internal.

% Harmful outcomes of actions that involve risk have one or more targets~(affected entities) and sources~(causing entity), which can be the main agent, other agents, or the environment. Risks could be due to two types of uncertainty: aleatoric (randomness in the environment) or epistemic (lack of experience or observation). The severity of the worst-case outcomes contemplated on the risks can be reversible, irreversible, or fatal to the agent. The update on the world from the harmful action's effects can either take place immediately or be delayed (risk is assessed over time). Lastly, repetition of the same action can result in varied outcomes over multiple occurrences such as breaking of an object by repeatedly hitting it. The following sections contain formal definitions for these ontological terms.

% \comment{Would be good to summarize all attributes and values of the ontology here before defining and describing each attribute.} Done

\subsection{External Attributes}

\subsubsection{Source}

\begin{definition}[Source]
    The \textbf{source} of a risk factor is the entity in the external world that is the main cause for the existence of the lesser utility transitions in the outcomes.
    
    A risk factor has a \textbf{primary} source if the causing entity is the agent itself, by lacking knowledge, having insufficient or imprecise interfaces, malfunctioning, etc.
    A risk factor has a \textbf{secondary} source if the causing entity is another acting entity in the world, such as other agents, humans, robots, etc.
    A risk factor has a \textbf{tertiary} source if the causing entity is some object or phenomenon of the environment itself. This includes the mechanics of how the environment works and how these are modeled in the MDP.
\end{definition}

Situations of risk in reinforcement learning are normally caused by one or more entities. The source attribute symbolizes this driving entity, responsible for the decreased utility in the states within the risk factor.

A risk factor with a primary source can be caused, for example, by a robotic agent that must maintain balance for itself, such as a Nao robot~\citep{garcia2020teaching}. Given that the agent can, willingly or unwillingly, cause harm to itself due to its own action---by letting itself lose balance and fall---it is the source of this risk factor.

An example of a risk factor with a secondary source is self-driving car scenarios, where an agent must consider that the actions and driving patterns of other vehicles on the road can affect its utility~\citep{yang2021cps}.

Most commonly in reinforcement learning, sources of risk factors are tertiary. Risk is caused by the physical laws or the mechanics of the environment itself, usually represented as increased or decreased rewards or costs, as well as by the existence of constraints or error states. For instance, the risk of having an agent fall into a hole can mainly be attributed to the environment as it features such dangers; thus constituting a tertiary source risk factor~\citep{ge2019safe}.

\subsubsection{Target}
\begin{definition}[Target] 
    The \textbf{target} of a risk factor is the main entity in the external world being affected should a transition in the outcome with lesser utility be chosen.
    
    A risk factor has a \textbf{primary} target if it affects the agent by hindering its progress toward completing its intended goal.
    A risk factor has a \textbf{secondary} target if it affects other acting entities, such as agents, humans, robots, etc.
    A risk factor has a \textbf{tertiary} target if it affects objects or phenomena in the environment itself, such as infrastructure, resources, etc.
\end{definition}

Reinforcement learning agents must not only complete their main objective but they are also expected to do so safely. This means protecting itself from danger, as well as human beings it interacts with, other agents, other living beings, objects and resources, and the environment itself.  If some entity important to the agent is affected by its action, this should reflect a loss in utility. Target is thus the entity that is most affected by a risk.

An example of a primary target is having the agent run out of fuel while driving a car, as it removes its ability to act and makes it impossible to accomplish its goal~\citep{simao2019safe}. Similarly, any risk outcome that may physically harm the agent is a primary target. A prime example of this is courses where the agent can crash against obstacles~\citep{maeda2021reconnaissance}.

If an action could cause harm or inconvenience to another person, the risk factor has a secondary target. In self-driving car domains, the actions of an agent might endanger the car's passengers and the passengers of other vehicles. In multi-agent environments, a risk factor may involve obstructing a cooperating agent, indirectly hindering our goal~\citep{bedi2021intermittent}.

Lastly, the actions of the agent may change the environment, constituting the tertiary target cases. In the bottle-carrying experiment, dropping bottles will pollute the environment, which in turn hinders the agent's ability to receive future rewards~\citep{vamplew2021potential}.

\subsubsection{Severity}
\begin{definition}[Severity]
    The \textbf{severity} of the risk factor refers to the degree to which the target is affected, should the transition with the least utility ever be visited.
    
    A risk factor has \textbf{no} severity if the MDP is theoretical, meaning that it does not represent a real-world scenario.
    A risk factor has \textbf{low} severity if its worst transition entails a minor inconvenience for the target. Depending on the target this means bumping or bothering a person, bumping or knocking objects, etc. Low severity usually entails no resources being expended as a consequence of the transition.
    A risk factor has \textbf{medium} severity if its worst transition entails temporarily changing the target. Depending on the target this means bruising a person or lightly damaging the agent’s vehicle or objects in the environment. Light severity may require some amount of resources to undo the change.
    A risk factor has \textbf{high} severity if its worst transition entails permanently changing the target. Depending on the target this means seriously injuring a person, or severely damaging the agent’s vehicle or other objects in the environment. High serenity means this change is impossible to completely undo or would require an exorbitant amount of resources to do so.
    A risk factor has \textbf{fatal} severity if its worst transition entails the complete cessation of functions of the target. Depending on the target this means the destruction of an agent’s vehicle, the death of a human, or the destruction of objects, resources, or phenomena in the environment.
\end{definition}

Severity is an assessment of how bad the worst possible result is in a risk factor. Choosing a severity to label specific risk factors will be very dependent on the domain and scenario. In many RL problems, we deal with computational simulations and no harm can be done to any real entity. However, many studies aim toward deploying agents in real-world circumstances. Moreover, different environments will feature various degrees to which agents could harm other entities and vice-versa. An agent that controls a robot may bump and knock over objects or people, but an agent that drives a card may cause a fatal accident. Each RL application is its own universe, and we merely provide some initial guidelines to which designers can label and evaluate how severe are the dangers in the world.

Purely theoretical scenarios involve no severity in the risk factors. Risky actions in grid world environments~\citep{lee2017constrained,pathak2018verification,prakash2019improving}, or in video games~\citep{saunders2017trial,jansen2020safe,garcia2022instance} often entail no real-world danger.

A scenario with low severity is having the agent leave broken bottles in the bottle-carrying experiment~\citep{vamplew2021potential}. If this happens, someone would have to pick the glass shards up. Arguably this severity could be higher if people frequented the path where the glass shards are.

Risk factors with medium severity include scenarios where there is some potential loss of resources due to the agent's actions, like in portfolio balancing experiments~\citep{bisi2022risk,lim2022dynamic} where an agent has to maximize profit. If the agent is in control of costly robotic equipment~\citep{garcia2020teaching,thomas2021safe}, having the agent lose control and crash can also be considered a medium-severity risk factor.

High-severity risk factors are characterized by irreversible damage or loss. This could be the bankruptcy of a business~\citep{garcia2019probabilistic}, or failure to deliver an important development project~\citep{tlili2021risks}. This would similarly cover heavy, irreversible injuries to human beings.

Risk factors with fatal severity have the death of one or more human beings as the worst-case scenario. Most often in reinforcement learning fatal risk factors are encountered in self-driving car scenarios~\citep{yang2021cps} or in the medicine domain~(keramati2020being).

\subsection{Internal Attributes}

\subsubsection{Uncertainty}

\begin{definition}[Uncertainty]
    The \textbf{uncertainty} of a risk factor refers to the nature of the agent’s knowledge about the outcomes. Given the agent’s lack of knowledge of the distribution of the MDP, all risk factors entail epistemic uncertainty: the uncertainty product of not knowing. However, some risk factors may also involve aleatoric uncertainty: the uncertainty product of randomness.
    
    A risk factor has \textbf{epistemic} uncertainty if the associated outcome is an exploration risk but not a domain risk.
    A risk factor has \textbf{epistemic+aleatoric} uncertainty if the associated outcome is both an exploration risk and a domain risk.
\end{definition}

By definition, the agent starts knowing nothing about the MDP. Thus, the agent has uncertainty over the results of the actions available. Over iterations of learning, the agent will approximate the MDP and its transitions. If the outcome of an action is however random, the agent also retains some uncertainty about the result. The uncertainty attribute of a risk factor determines whether an agent can avoid danger once it fully understands the world, or whether some actions have a chance of backfiring.

Epistemic uncertainty is the dangers or unfavorable outcomes for the agent that can be avoided once enough knowledge is accumulated. In grid world~(prakash2019improving,wachi2020safe) and robotic control~\citep{verbist2018actor,liu2022safe,yang2022safety,cowen2022samba} environments, non-moving obstacles, and hazards are common instances of epistemic risk factors. Once the agent learns their position, they can be completely avoided.

On the other hand, some risk factors may contain both epistemic and aleatoric uncertainty. In these cases, certain state-action combinations will have some degree of uncertainty in their results, as the utility of the resulting state may vary. If to the previous examples, we add movement to the obstacles~\citep{yang2022safety} or slippery movement near the grid world hazards~\citep{maeda2021reconnaissance,turchetta2020safe}; those now have both types of uncertainty.

\subsubsection{Interference}
\begin{definition}[Interference]
    The \textbf{interference} of a risk factor refers to the degree to which the agent’s goal is affected, should the transition with the least utility ever be visited.
    
    Let $G \subset S$ be a set of goal states.
    A risk factor has \textbf{reversible} interference if the agent can reach any state in $G$ without any loss in its final reward. For example, if an agent can backtrack decisions so it can collect the reward for visiting some state it missed, that risk factor has reversible interference.
    A risk factor has \textbf{irreversible} interference if the agent can reach any state in $G$ with some loss in its final reward.
    A risk factor has \textbf{blocking} interference if the agent can no longer reach any state in $G$.
\end{definition}

Given no hindering obstacles or hazards, stochasticity in the movement of an agent~\citep{lee2017constrained,chow2018lyapunov,serrano2020safe} is a reversible risk factor as the agent can take action to steer back into its preferred course. Actions that have a cost, punishment, or violate a constraint are types of irreversible risk. For instance, in a robotic control environment where we want to preserve the integrity of the agent's vehicle, high-torque movements are an irreversible risk factor that might damage the robot~\citep{bisi2022risk}. Blocking risk factors include those that restart the iteration process, such as falling into a hole in a grid world environment~\citep{ge2019safe,serrano2020safe,maeda2021reconnaissance} or crashing on a self-driving car simulation~\citep{cheng2019end,alshiekh2018safe}.

\subsubsection{Frequency}
\begin{definition}[Frequency]
    The \textbf{frequency} of a risk factor refers to the likelihood that transitions on the outcome are visited.
    Frequency depends on the average probability to visit a transition from the outcome, calculated from each applicable state-action pair. Let $Z$ be the set of all state-action pairs that have at least one transition in $O$ such that $(s,a)\in Z \leftrightarrow \exists_{s_{2} \in S} (s,a,s2) \in O$. The frequency of a risk factor can be calculated as:
    $$
        p = \frac{1}{|Z|}\sum_{(s,a) \in Z}  \sum_{(s,a,s_{2}) \in O} T(s,a,s_{2})
    $$

    A risk factor has \textbf{guaranteed} frequency when $p=1$.
    A risk factor has \textbf{common} frequency when $0.5 \leq p < 1$.
    A risk factor has \textbf{sporadic} frequency when $0.25 \leq p < 0.5$.
    A risk factor has \textbf{uncommon} frequency when $0.1 \leq p < 0.25$.
    A risk factor has \textbf{rare} frequency when $0.01 \leq p < 0.1$.
    A risk factor has \textbf{very rare} frequency when $p < 0.01$.
\end{definition}

While the exact frequency value can be calculated for a concrete MDP and outcome, in most scenarios defining both is impractical. For risk characterization, we recommend doing an approximation of this probability. Similarly, the exact probabilities proposed here can be adjusted to the specific application domain, as in some cases the probabilities of undesirable outcomes can be higher or lower.

The way frequency is calculated is using the probability of visiting a transition in the risk factor, for every state-action pair in the risk factor. As such, the actual value of $p$ depends on how is the risk factor defined. For example, if we consider an environment where an agent has a 66\% chance to slip to a random direction when walking in an intended direction---and a 33\% to walk in the intended direction---the risk factor will have a common frequency. A similar environment would be a stock or asset trading environment, where the prices change every timestep. In this case, we could consider every transition to be a risk factor as there is no `intended' direction, which would have a guaranteed frequency. The difference between these two cases is the existence of intent with an action and highlights that this property is highly dependent on how are risk factors defined.

\subsubsection{Timing}
\begin{definition}[Timing]
    The \textbf{timing} of a risk factor refers to whether the difference in utility between transitions is observable in the next time step, or if it is delayed to future time steps.

    A risk factor has \textbf{immediate} timing if it is a risk factor under the immediate utility function $U(t) = R(t)$. For domain risks, the following must be true

    $$
        \forall_{(s_{1},a,s_{2}) \in O} \exists_{s_{3} \in S}: (s_{1},a,s_{2}) \notin O \land R(s_{1},a,s_{2}) < R(s_{1},a,s_{3}).
    $$

    For exploration risks that are not domain risks, the following must be true:
    $$
        \forall_{(s_{1},a,s_{2}) \in O} : K(s_{1},a) < k \land R(s_{1},a,s_{2}) < u.
    $$

    Otherwise, the risk factor has \textbf{delayed} timing.
\end{definition}

The idea behind the timing property is to categorize risks depending on whether an action will have immediate consequences, or if these will be observable later. Most ``death states'' in RL conform risk with immediate timing, as the agent loses its ability to act on the next timestep. Delayed risk factors include time limits to complete a task~\citep{zhang2021safe}, changes to the environment with future effects~\citep{vamplew2021potential}, or actions with long-term consequences such as buying stocks~\citep{zhang2022safe}.

\section{Mapping Results}
\label{sec:resu}

\subsection{RQ1 -- Ontological Mapping}
\label{sec:resu-rq1}

\begin{figure}
    \centering
    \includegraphics[width=0.9\linewidth]{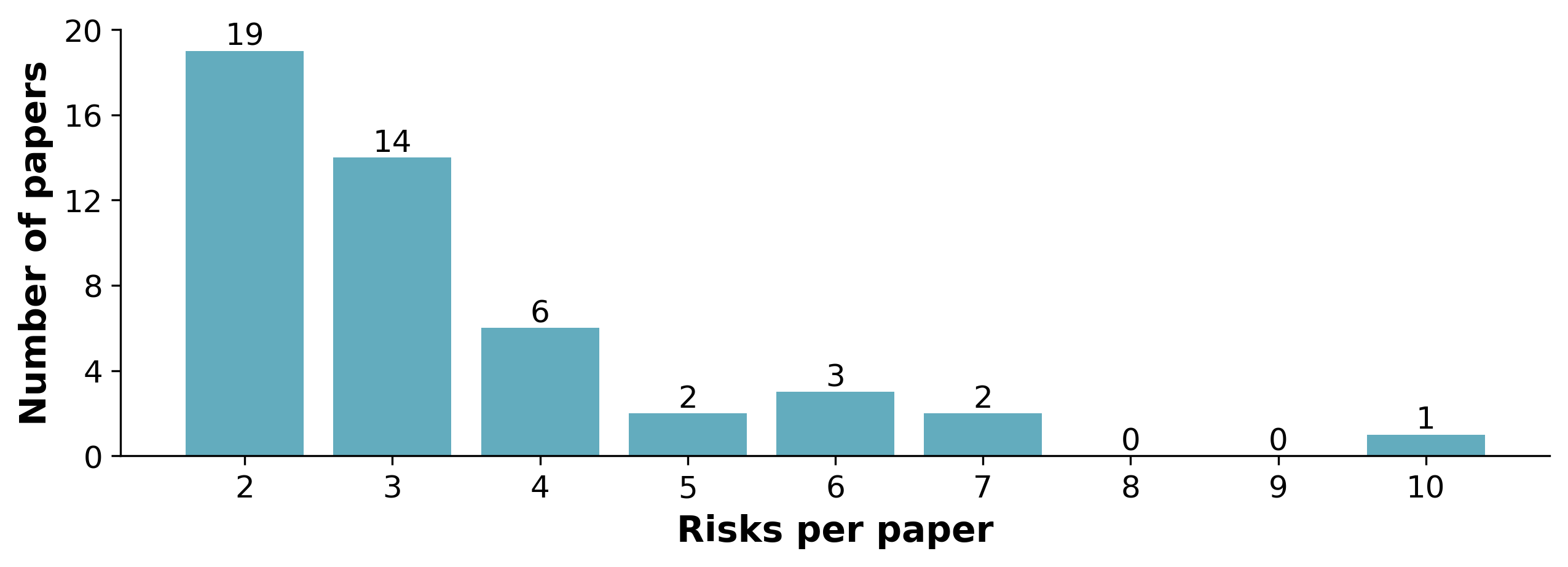}
    \caption{Distribution of the number of risks identified per paper.}
    \label{fig:risk-histo}
\end{figure}

\ref{rq:r1} aims to characterize the risks that are studied in safe RL. 
We are particularly interested in identifying, characterizing and labeling risk factors~(Section~\ref{sec:chara}), and assessing how frequently these have been studied. In total, we identified 178 risk factors in the \numpapers covered papers. Figure~\ref{fig:risk-histo} shows a histogram of the identified risks per paper. 
% Counts of risk in papers are plotted on the x-axis whereas the number of papers are outlined on the y-axis. 
A safe RL paper covers 2.47 risk factors on average, and the median value similarly was 2.

\subsubsection{Attribute Mapping}

\begin{figure}
    \centering
    \includegraphics[width=\linewidth]{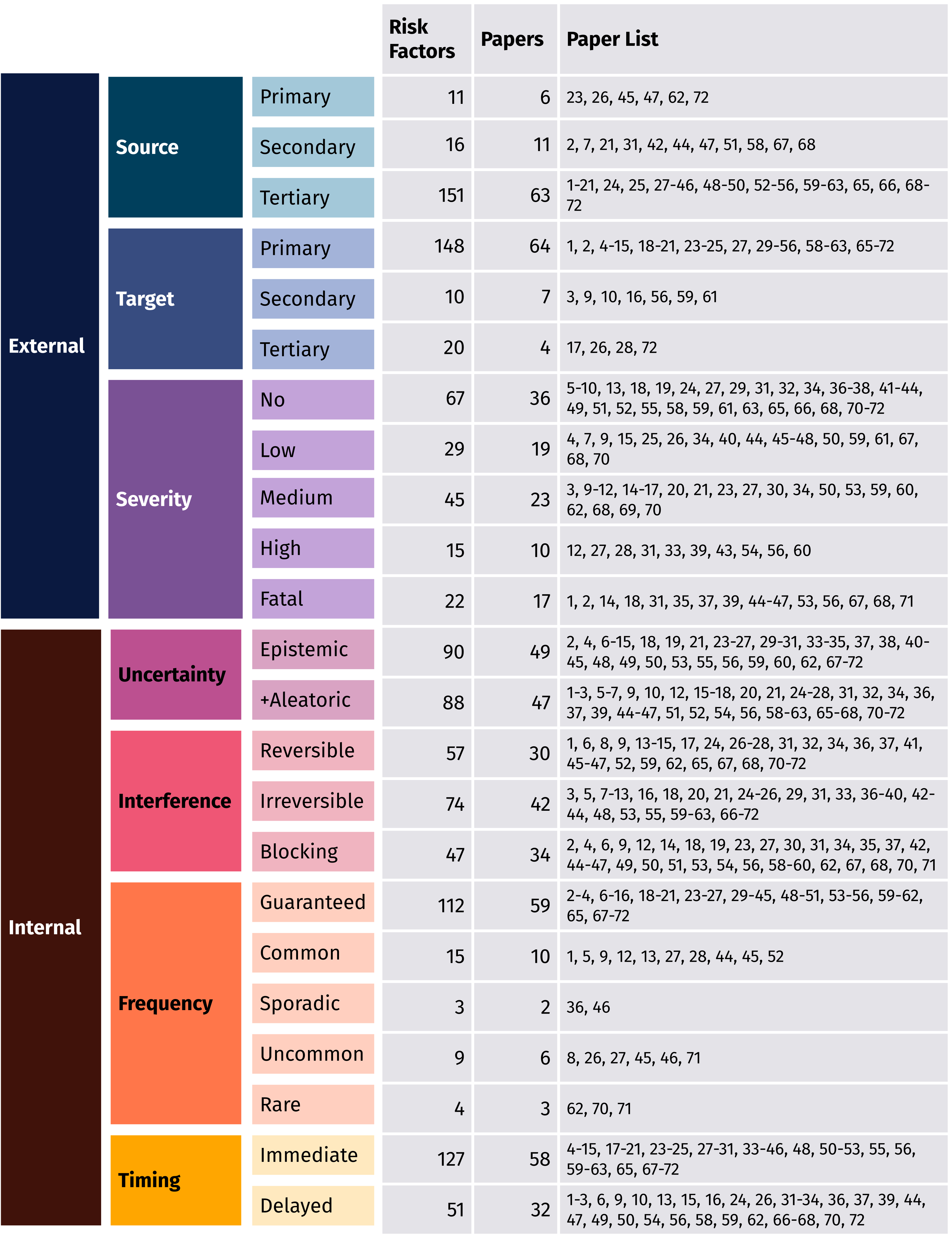}
    \caption{Breakdown and counts of risk factor characteristics.}
    \label{fig:risk-onto}
\end{figure}

We characterized the identified risk factors in our proposed attribute system. Figure~\ref{fig:risk-onto} shows the breakdown of the risk attributes and the number of papers that have at least one risk with the respective value.

\textbf{Source}
The grand majority of identified risk factors dealt with tertiary sources, with 151 factors and 63 papers. Risk factors with secondary and primary sources are covered in the RL literature but within more specific applications.

\textbf{Target} 
On a trend similar to the source attribute, the most common type of target for risk factors was primary, with 148 occurrences across 64 studies. Most of the studied risks in the literature primarily threatened the agent. Tertiary and secondary targeting risks are still present in some studies but remain a relatively uncovered area.

\textbf{Severity}
The mode in risk factor severity was no real-world repercussions, with 67 factors and 36 papers. This is due to the related experiment's nature as an abstract, proof of concept example. The remaining 111 risk factors can however be associated with some real-life threat. Most commonly in these studies were low and medium severity levels. However in certain applications such as robotics, self-driving cars, and patient treatment; the risks we identified carry a heavy meaning should these systems be ever used. There is thus a significant amount of ongoing research on higher-severity risks.

\textbf{Uncertainty}
Roughly half of the risk factors had epistemic uncertainty only, and the other half also expressed aleatoric uncertainty. This highlights the importance of safety during the exploration of the environment, contrary to only considering the aspect of safety during the creation of the optimal policy. Roughly half of the risks an agent faces can be mitigated with safer exploration, according to these results.

\textbf{Interference}
Most commonly, studies in reinforcement learning faced risk factors that had irreversible consequences, but that allow the agent to continue exploring the environment. A total of 74 factors and 42 papers consider irreversible risk. If to this number we add blocking~(makes goal impossible) risk factors, we get 121 factors. This again shows that the risks a reinforcement agent faces have dire consequences in real-life environments, and highlights the importance of safe exploration policies. Reversible risks were still abundant in the literature and remain an important part of safe RL.

\textbf{Frequency}
Not shown in figure~\ref{fig:risk-onto} are the 35 risks that we could not attribute a specific frequency, due to a lack of information on the specifics of the experiment or excessive complexity of the underlying MDP. We similarly did not identify any risk factors with very rare frequency. The overall trend suggests that safe RL research focuses on the dangers that are very likely to happen or are guaranteed to happen under certain circumstances.

\textbf{Timing}
Most risk factors in the literature had an immediate negative impact on the agent's perceived utility, with 127 factors and 58 studies. Delayed risk factors were still covered in the safe reinforcement learning literature, but remain an interesting area for future research.

\subsubsection{Risk Factor Types}

Once we identified and characterized the risk factors in the safe RL literature, we grouped them into types. A risk factor type represents a general representation of a real-life risk as an element in the MDP. Because of these, risk factors under a type have similar characteristics. Table~\ref{tab:riskf} shows the 13 proposed risk factor types, the number of risk factors labeled, and the most common (mode) value for each characteristic. We abbreviated the values in the table using the first 3 letters of their name.

\begin{table}[ht]
\centering
\caption{Identified risk factor types, occurrences, and most common attribute value.}
\label{tab:riskf}
\begin{tabular}{@{}lrlllllll@{}}
\toprule
\textbf{Risk Factor} & \multicolumn{1}{l}{\textbf{\#}} & \textbf{Src.} & \textbf{Tgt.} & \textbf{Sev.} & \textbf{Unc.} & \textbf{Int.} & \textbf{Frq.} & \textbf{Tim.} \\ \midrule
Hazard State         & 36                             & Ter             & Pri             & Med           & Epi           & Irr           & Gua           & Imm           \\
Death State          & 32                             & Ter             & Pri             & Fat           & Epi           & Blo           & Gua           & Imm           \\
Random Transition    & 22                             & Ter             & Pri             & No            & E+A           & Rev           & Gua           & Imm           \\
Constraint           & 20                             & Ter             & Pri             & No            & Epi           & Irr           & Gua           & Imm           \\
Cost                 & 16                             & Ter             & Pri             & No            & Epi           & Irr           & Gua           & Imm           \\
Domain-Specific      & 14                             & Ter             & Ter             & Med           & E+A           & Rev           & Unk           & Imm           \\
Variance             & 10                             & Ter             & Pri             & No            & E+A           & Irr           & Gua           & Imm           \\
Random Start         & 6                              & Ter             & Pri             & No            & E+A           & Rev           & Gua           & Del           \\
Danger Sense         & 6                              & Ter             & Pri             & Fat           & Epi           & Blo           & Gua           & Imm           \\
Pollution            & 6                              & Pri             & Ter             & Low           & Epi           & Irr           & Gua           & Del           \\
Adversary            & 5                              & Sec             & Pri             & No            & E+A           & Blo           & Unk           & Del           \\
Random Objective     & 3                              & Ter             & Pri             & No            & E+A           & Rev           & Gua           & Del           \\
State Obfuscation    & 2                              & Pri             & Pri             & Low           & E+A           & Rev           & Com           & Del           \\ \bottomrule
\end{tabular}
\end{table}

Hazard states were the type with the most identified risk factors: a total of 36. These represent transitions that result in a set of states that are considered unsafe for the agent to be in and that should be avoided. Normally these are inflicted by the environment (tertiary source) and have a direct negative consequence for the agent (primary target). Hazard states however are not a complete hindrance to the agent's goal, and studies commonly count the frequency to which agents visit these states. As such, hazard states often have medium severity and irreversible interference. Hazard states are often deterministic (guaranteed frequency), immediately noticeable for the agent (immediate timing), and can be avoided with enough knowledge (epistemic uncertainty).

The second most frequent risk factor type was death states, with 32 occurrences. These are very similar to hazard states, with the difference that entering a death state means the agent fails at its task. Because of these, it shares the most common attribute values with hazard state, except these have fatal severity and blocking interference.

Random transitions represent risk factors associated with the execution of another action instead of the one intended by the agent. In this type of risk factor, there is an intended transition for every state-action pair, and at least one other transition with the same origin state-action pair but a different resulting state. In other cases, there is no intended transition, but the action will still exhibit a random outcome (guaranteed). As such, this phenomenon is inherently aleatoric. By themselves, these pose no danger to the agent (no severity, reversible, immediate) but they can interact with other risk factors. For example, having both random transitions and death states in the same MDP makes the death state have epistemic and aleatoric uncertainty. Random transitions have a primary target, but the source can be attributed either to the environment (slippery floor) or the agent itself (imprecise controls).

Constraints are external, user-defined defined rules (tertiary) that affect the agent's interaction with the MDP (primary). Ideally, a trained agent will learn and work within the imposed constraints (epistemic), and their violation incurs a penalty (guaranteed, immediate, irreversible).

Very similar to constraints, costs are an irreversible risk factor associated with specific actions. An agent trained in an environment with costs will try to minimize or completely avoid any actions with costs but can elect to take costly actions if needed. The main difference with constraints is that costs often had less severity (though note the mode for both was no severity).

Domain-specific risks are a blanket type that contained all risk factors that were unique enough to not fit into any of the other categories. This was comprised of risks specific to software project management, which are factors that an agent must reason about. They are related to factors of the environment that the agent can try to control, and affect the environment in turn.

The variance type refers to risk factors that represent variance in the immediate reward of an agent (primary target) for selecting an action. In many of the MDPs in safe RL literature, actions yield rewards (tertiary source) from a probability distribution (aleatoric) instead of a set amount, however ending in the same resulting state. In most cases the agent cannot repeat the action to `farm' the reward, meaning the potential win or loss of reward is irreversible. The obtained reward is immediately observed, and the guaranteed frequency of the risk is guaranteed (the agent is sure to receive a random reward as there is no `intended' value).

Random starts are risk factors that have the agent start on one of a set of states at the beginning of a learning episode. By nature, these are aleatoric and usually can be reversed through a sequence of actions. Because the agent does not receive any reward when starting an episode, random starts are considered delayed risks.

In certain applications, agents have a `danger sense' that allows them to know how unsafe the current state is. For example, an agent may have access to the distance of another car or agent, the distance to the closest dangerous area, or how much it is straying from its designated path. Danger sense risks rely on another risk factor that represents a threat, such as hazard or death states. Thus, while not directly harmful to the agent itself, danger sense gives an additional bit of information the agent can reason about to try to avoid risk.

Pollution-type risks are permanent (irreversible) changes that an agent (primary source) inflicts into the environment (tertiary target). In most of the observed instances, certain actions are guaranteed to cause harm to the environment (guaranteed, epistemic). While not immediately threatening the agent's mission, the damage to the environment might hinder the agent's ability to reach its goal down the line (delayed timing).

Applications like games or multi-agent environments often feature an adversary: another acting entity (secondary source) that has a goal that is opposite to the agent's own (primary target). The adversary will, at every opportunity, take actions that will impede the agent's goal (blocking).

Random objectives are risk factors that randomly select the goal set of states at the beginning of every episode. The set of goal states is hidden from the agent, meaning they must discover them through trial and error; alternatively, this information could be available and observable from the state.

State obfuscation refers to the risk of having an agent in a partially observable MDP. This means the agent is unaware of what is its current state, and must instead consider possible states it is in and try to deduct this information by observing the outcome of actions.

\subsection{RQ2 -- Application Domains}
\label{sec:resu-rq2}

\begin{figure}
    \centering
    \includegraphics[width=0.9\linewidth]{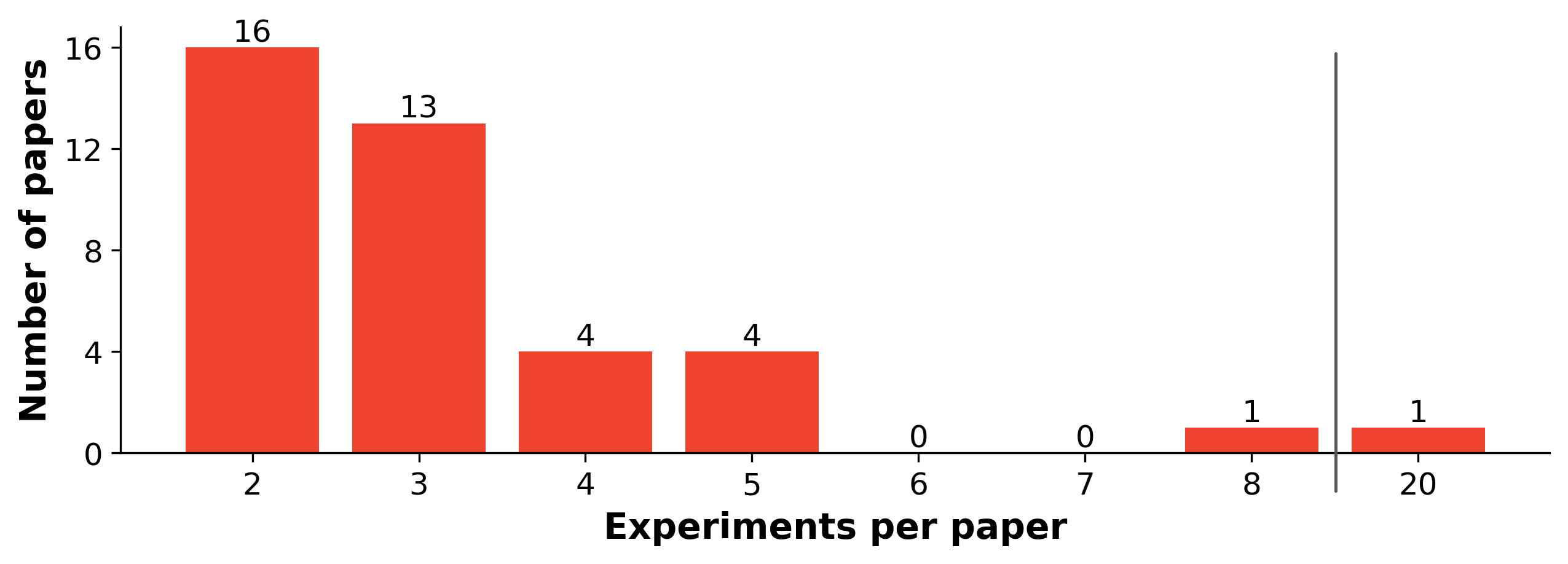}
    \caption{Distribution of the number of experiments identified per paper.}
    \label{fig:risk-histo2}
\end{figure}

\ref{rq:r2} aims to identify the application domains studied in safe RL. Our aim is to identify trends in research domains, particularly if studies involve domain-specific or domain-free techniques. We identified a total of 168 experiments in the surveyed papers. Figure~\ref{fig:risk-histo2} shows a histogram of the number of experiments identified per paper, with an average of 2.3 and a median of 2 experiments.

We classified these experiments into 5 domains: toy problems, engineering, economics, medicine, and computer science. Toy problems refer to small example cases, simple MDPs, grid-based and continuous-walking scenarios, and simple games. Engineering mainly contemplates model and robot control problems. Economics in safe RL often contemplates portfolio balancing and trading experiments. Medicine similarly encompasses health-related experiments like drug administration. Lastly, we classify more complex cases of MDPs and games (such as arcade and Atari games), or CS-related scenarios such as software project management under computer science.

\begin{figure}
    \centering
    \includegraphics[width=0.85\linewidth]{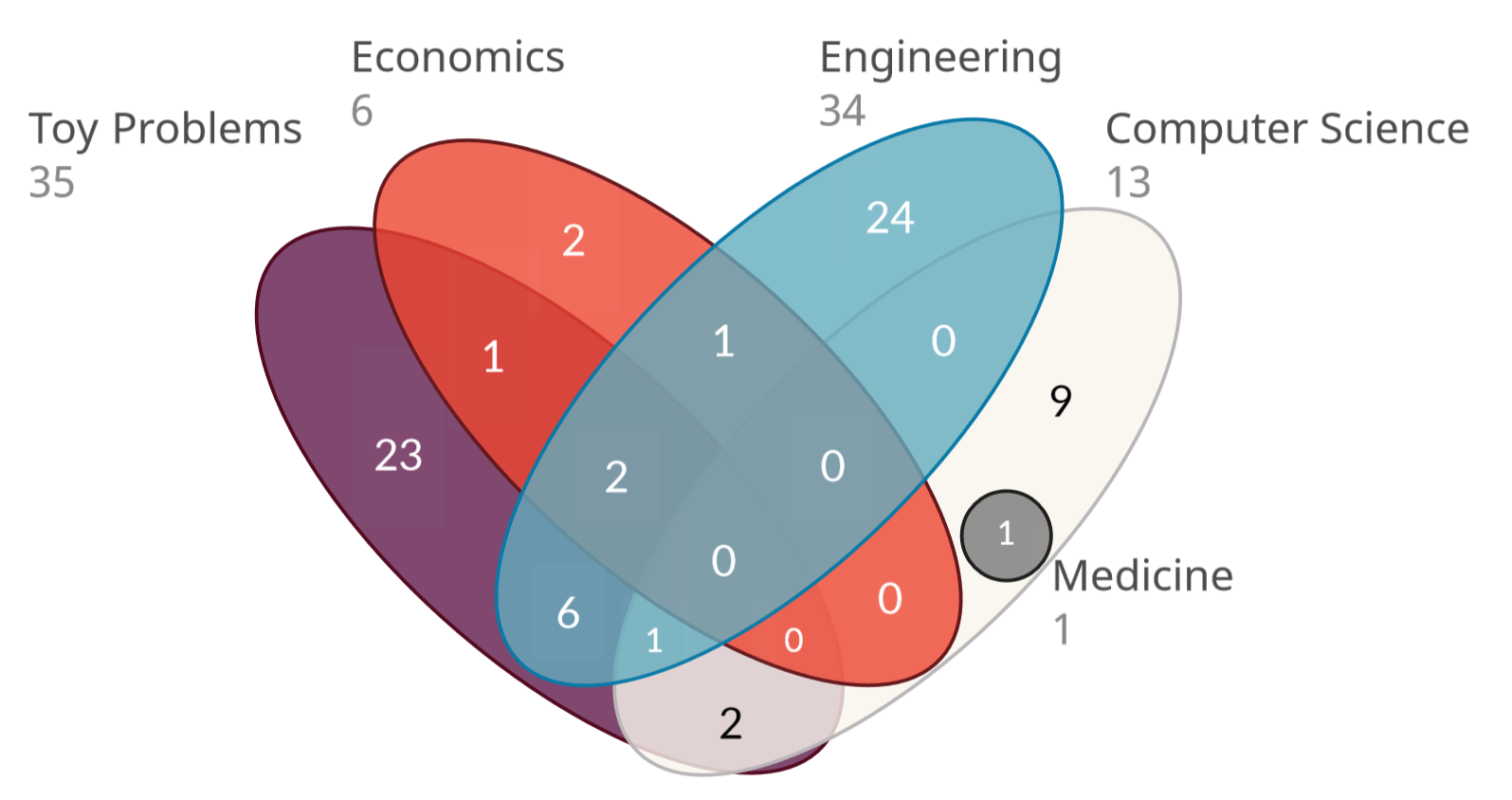}
    \caption{Number of papers covered by domain and their overlap.}
    \label{fig:domain-venn}
\end{figure}

Figure~\ref{fig:domain-venn} shows the number of papers covered by each domain and also shows how many papers exist in certain overlaps of the five core disciplines focused on: toy problems, economics, engineering, computer science, and medicine.

Most frequently---58 out of 72 surveyed papers---, studies were concerned with experimenting on just one domain. Studies thus have a tendency of proposing techniques specialized towards one domain. A common trend in the surveyed papers was to present experiments with varying degrees of complexity. Most simple experiments fell under toy problems, so almost all intersections in Figure~\ref{fig:domain-venn} are with the toy problem domain. The domain intersection with most studies was toy problems and engineering, with 9 total papers. Such cases comprised classical RL scenarios: grid walking problems and robotic or model controls. Computer science+toy problems, economics+engineering, and toy problems+economics were similarly related with 3 papers each overall. Three and four-domain intersections were seldom encountered, with 3 studies in total. Intersections of computer science with non-toy domains, as well as intersections with medicine, were barely studied. There is thus a research gap in safe RL algorithms that can be applied to multiple domains.

\begin{figure}
    \centering  \includegraphics[width=0.95\linewidth]{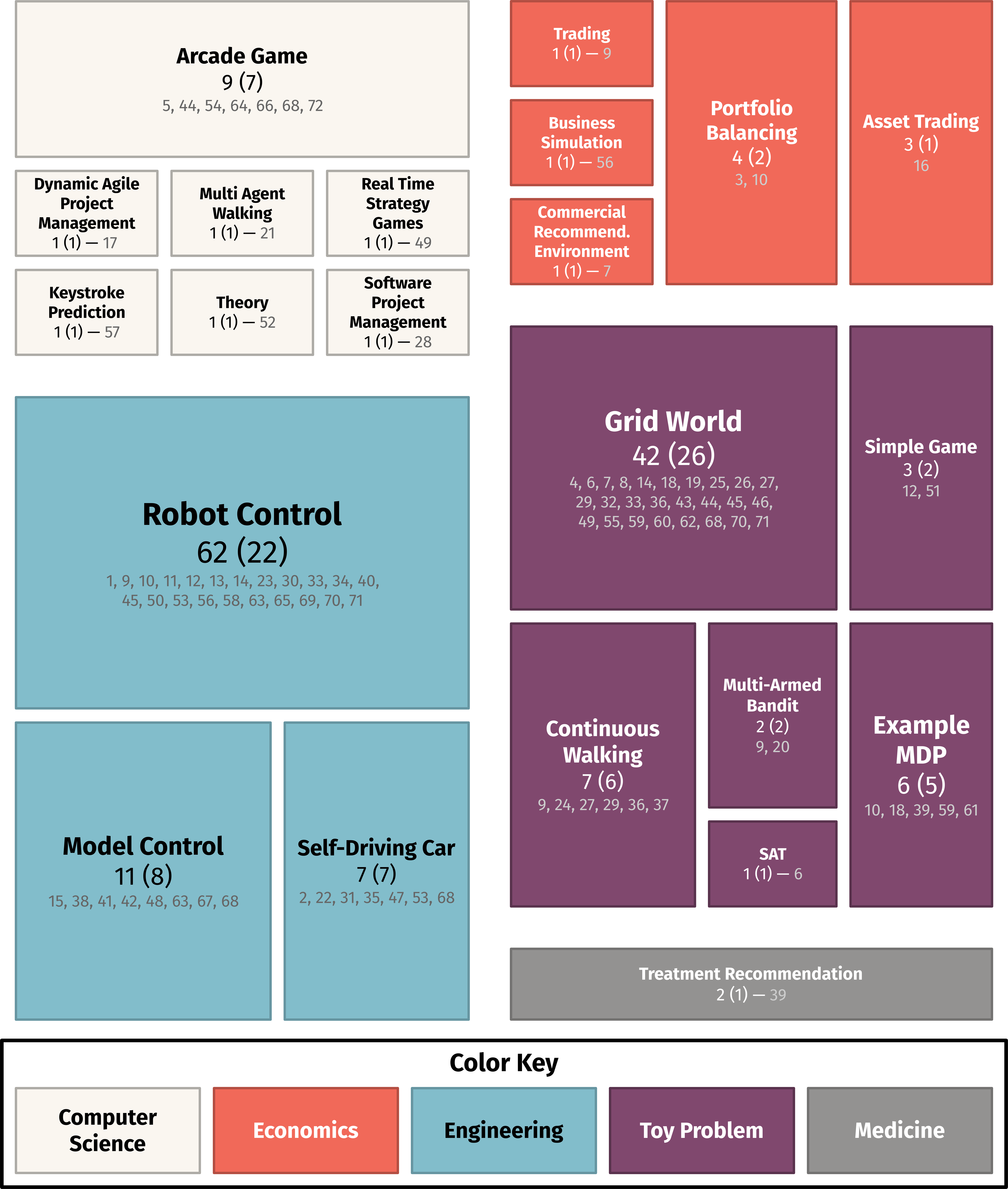}
    \caption{Experiment types grouped by domain, occurrences, (number of papers), and list of papers.}
    \label{fig:experiment-counts}
\end{figure}

Similarly, Figure~\ref{fig:experiment-counts} details the number of experiments and (papers) of each experiment type, grouped by domain. Instances with a number were encountered only once. Robot control was the most frequent experiment, with a total of 62 instances across 22 papers. While common, experiments under this category were diverse---ranging from piloting a NAO robot~\citep{garcia2020teaching}, to piloting a helicopter~\citep{pan2018efficient,garcia2019probabilistic}, and the use of OpenAI Gym~\citep{cowen2022samba,thomas2021safe,luo2021learning,serrano2020safe,stooke2020responsive,cheng2019end} and MuJoCo~\citep{bisi2022risk,liu2022safe,zhang2022safe} environments. Grid world or grid walking scenarios were the second most common, with 42 instances across 26 papers, featuring an agent that maneuvers in a 2-dimensional grid to complete some objective. While most studies proposed unique grid world experiments, there are two scenarios that are common in the safe RL literature---cliff world~\citep{xuan2022sem,simao2019safe,serrano2020safe,ge2019safe,lee2017constrained} and frozen lake~\citep{maeda2021reconnaissance,van2021no,turchetta2020safe}. Model control experiments have the agent learn to control or optimize a task, such as liquid tank balance~\citep{zanon2022stability,alshiekh2018safe}. Arcade games were the most studied scenario in the computer science category, mainly Atari games~\citep{garcia2022instance,dabney2018implicit,saunders2017trial,dilokthanakul2018deep}. Continuous walking, self-driving cars, portfolio balancing, and asset trading are categories of studies we encountered multiple times, yet seem to have the potential for further research.

Table~\ref{tab:risk-domain} shows the intersection between risk factor types and application domains, except the medicine domain. Risk factors in the toy problem domain follow the global trend, emphasizing hazard and death state, random transition, cost, and constraint risks. Engineering problems have a strong focus on three risk factors: hazard states, death states, and constraints. Economics studies risk is almost exclusively comprised of random transitions, whereas computer science studies similarly focus on domain-specific risks. Lastly, there were two risk factors in the medicine domain, both related to the variance of reward over time. These trends seem to show little compatibility with the studied risks across domains.

\begin{table}[t]
\caption{Risk factor type counts per application domain.}
\label{tab:risk-domain}
\begin{tabular}{@{}lrrrr@{}}
\toprule
\textbf{}            & \multicolumn{1}{l}{Toy Problem} & \multicolumn{1}{l}{Engineering} & \multicolumn{1}{l}{Economics} & \multicolumn{1}{l}{Computer Science} \\ \midrule
Hazard State         & 18                              & 15                              & 0                             & 3                                    \\
Death State          & 12                              & 16                              & 1                             & 3                                    \\
Random Transition    & 10                              & 5                               & 5                             & 2                                    \\
Constraint           & 7                               & 11                              & 1                             & 1                                    \\
Cost                 & 11                              & 4                               & 0                             & 1                                    \\
Domain-Specific Risk & 0                               & 0                               & 0                             & 14                                   \\
Variance             & 5                               & 2                               & 0                             & 1                                    \\
Random Start         & 4                               & 2                               & 0                             & 0                                    \\
Danger Sense         & 2                               & 3                               & 0                             & 1                                    \\
Pollution            & 5                               & 0                               & 0                             & 1                                    \\
Adversary            & 1                               & 4                               & 0                             & 0                                    \\
Random Objective     & 3                               & 0                               & 0                             & 0                                    \\
State Obfuscation    & 0                               & 1                               & 0                             & 1                                    \\ \bottomrule
\end{tabular}
\end{table}

\subsection{RQ3 -- Representation and Techniques}
\label{sec:resu-rq3}

\begin{figure}[ht]
    \centering
    \includegraphics[width=\linewidth]{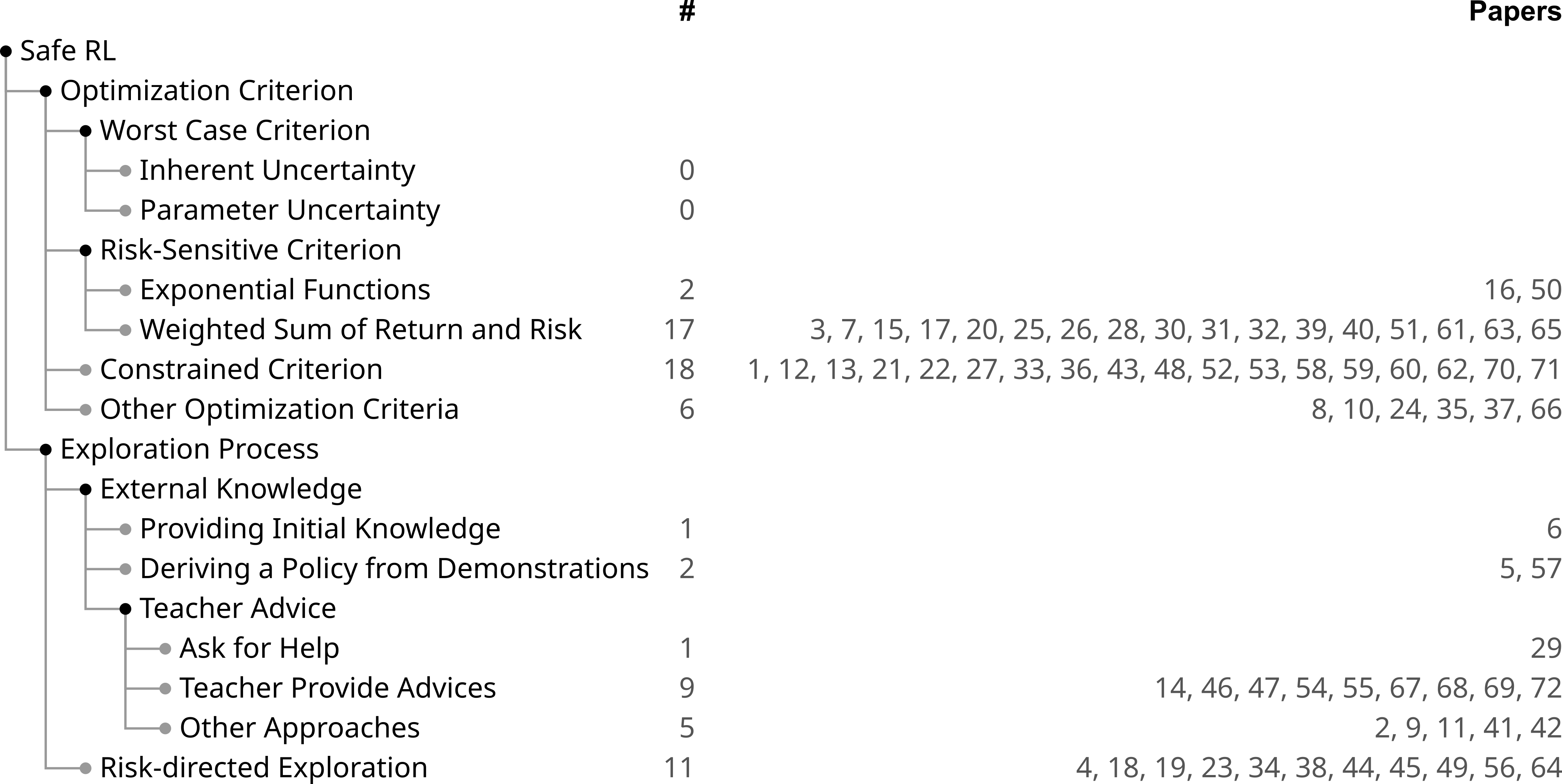}
    \caption{Mapping of papers to safe RL taxonomy.}
    \label{fig:taxonomy-counts}
\end{figure}

\ref{rq:r3} aims to identify the risk representation and techniques studied in safe RL. To this end, we classified the surveyed papers under the safe RL risk taxonomy~\citep{garcia2015comprehensive}. Figure~\ref{fig:taxonomy-counts} maps the number of papers to this taxonomy. In general, the safe RL literature favors optimization that considers safety in the long run rather than exploration that minimizes safety violations. Constrained criteria~\citep{cheng2019end} was the most common type of approach in safe RL, with 18 studies. For risk-sensitive criterion, most papers utilized a weighted sum of return and risk~\citep{adel2021multi}, yet a couple of others employed exponential functions~\citep{park2022uncertainty}. Other types of optimization criteria were commonly used~\citep{cohen2021model}, but we did not identify any studies that utilized worst case criterion---neither inherent or parameter uncertainty.  Papers on the branch of exploration favored the use of teacher advice approaches, mainly the use of techniques that have the teacher automatically provide advice~\citep{xuan2022sem}. There is still research in teaching approaches where the learner asks for help~\citep{wachi2021safe}, or others where the relationship is not clear~\citep{liu2022safe}. Risk-directed exploration approaches are relatively popular in safe RL~\citep{simao2021alwayssafe}, as it is a common way to try to train an agent while having it avoid any catastrophic states. While we encountered providing initial knowledge~\citep{hansmeier2022integrating} and deriving a policy from demonstration~\citep{huang2019risk} type techniques, these were in the minority.

Certain types of safe RL approaches were more common in certain domains than others. Table~\ref{tab:domain-taxo} shows the number of papers per risk taxonomy entry and application domain. Toy problems mostly favored constrained criteria, weighted sums of return and risk, and risk-directed exploration. However, this might be because many studies included the toy domain as a way to test their techniques, and not because they might be better at solving these types of problems. Engineering follows a similar trend to toy problems, with the addition of teacher advice approaches as a more commonly studied combination. Economic and medicine-oriented experiments instead favor optimization criteria, but the sample size for these areas is too small to establish a clear research pattern. Computer science instead favors exploration techniques and weighted risk and return. The main research pattern in this data is that toy problems and engineering commonly employ constrained criteria, while the other domains don't necessarily have a preference.

\begin{table}[t]
\centering
\small
\caption{Number of papers per risk taxonomy entry and application domain.}
\label{tab:domain-taxo}
\begin{tabular}{@{}lrrrrr@{}}
\toprule
 & \multicolumn{1}{l}{Toy} & \multicolumn{1}{l}{Econ.} & \multicolumn{1}{l}{Engi.} & \multicolumn{1}{l}{Comp.} & \multicolumn{1}{l}{Medi.} \\ \midrule
Exponential & 0 & 1 & 1 & 0 & 0 \\
Weighted R\&R & 8 & 2 & 6 & 4 & 1 \\
Constrained & 10 & 0 & 10 & 2 & 0 \\
Other Opt. & 4 & 1 & 2 & 1 & 0 \\
Initial Knowledge & 1 & 0 & 0 & 0 & 0 \\
Demonstrations & 0 & 0 & 0 & 2 & 0 \\
Ask For Help & 1 & 0 & 0 & 0 & 0 \\
Teacher Advice & 4 & 0 & 5 & 3 & 0 \\
Other Approaches & 1 & 1 & 5 & 0 & 0 \\
Risk Exploration & 6 & 1 & 5 & 3 & 0 \\ \bottomrule
\end{tabular}
\end{table}

\section{Discussion}
\label{sec:disc}

In this section, we discuss the implications of the results we uncovered on the mapping study. Our aim with this discussion is to highlight areas of future work and spark creativity in new research on safe RL.

\subsection{Accounting Risk in Experiments}
Preparing and designing a reinforcement learning experiment requires modeling a real-world problem---simplifying complex phenomena into a more manageable representation. However, by doing so, a designer could be discarding important factors; and a reinforcement learning agent could, when deployed back in the real-world scenario, cause harm or some otherwise undesirable results.

We strongly endorse that researchers and practitioners in reinforcement learning do a risk profiling process as they design their experiments. By doing so, they are validating that their model of the world contains all relevant risks so that an agent that is deployed will minimize unwanted behavior. Moreover, having a profile of the risks in the environment aids in the selection of the particular RL algorithm, design of rewards and optimization function, and exploration strategy.

As we are proposing these definitions of risk and its characterization, we do not have yet a sound methodology for the risk profiling process for novel RL problems. We suggest using the following list of questions to identify potential risk factors in an environment:
\begin{enumerate}
    \item What is the goal of the agent? What should it learn to do?
    \item What stands in the way of the agent when it learns this task?
    \item Once it has enough knowledge, what stands in the way of the agent when it executes this task?    
    \item What are the situations the agent should avoid at all costs? Are these more important than completing the goal?
    \item What are the situations the agent should avoid if possible?
    \item Are there any acting entities in the environment?
    \begin{itemize}
        \item Are they helpful for the task? Should the agent protect them? 
        \item Are they harmful to the task? Should the agent plan around them? 
    \end{itemize}
    \item Can the agent change the environment? Can that change help or hinder its ability to reach the goal?
    \item Is the agent able to see reality correctly? Or could its sensors have some degree of imprecision?
    \item Is the agent able to execute its actions as intended? Or could its actuators have some degree of imprecision?
    \item Will the initial and final conditions ever be different?
    \item Is there any way the agent could perform the task, but better? What changes?
\end{enumerate}

By answering these questions, a designer will have a list of potential risk factors. We suggest then characterizing them according to our proposed attributes. It is likely that not all risk factors can be accurately represented in the simulation, so the designer might have to pick a subset of important factors. Lastly, the risk profiling process is iterative: risks that were previously not considered might rise at any point during the reinforcement learning process.

\subsection{The Case of Negative Rewards}
Traditional reinforcement learning uses rewards to endorse or punish behaviors for an agent. For example, agents are normally given a big reward for completing a task, a big punishment for failing, and small rewards for achieving progress toward the goal. The reward shaping task is a burden to the designer of the MDP, however, as they have to test that the agent will learn the intended behavior from the chosen rewards and optimization function. It is not uncommon for RL agents to engage in `reward farming' behavior in which they fail to complete the task and instead abuse design flaws in the MDP that allow them to collect a potentially infinite reward.

The existing representations of risk in RL propose a solution to this problem. Constraints can limit the maximum number of actions an agent can take. Costs can discourage the agent from taking certain actions over and over. Costs and constraints often use mechanisms other than reward to provide this feedback to the agent: the cost function. This adds the problem that the utility function of the agent must now accommodate for reward and risk, and how these can be traded off. And again, this burden falls to the designer.

Current research in multi-objective reinforcement learning sheds light on this problem. Instead of manually designing rewards and utility functions, multi-objective RL proposes a framework for finding a preference as the agent learns~\citep{hayes2022practical}. This relieves the designer from the trial-and-error process of reward shaping and instead emphasizes the selection of behavior. With our proposed view of studies having multiple sources of risk to consider, multi-objective RL is a potential sub-area to consider for future research.

\subsection{Scarcity of Multi-Domain Studies}
As highlighted in the second research question, the majority of safe RL studies emphasize one application domain to perform experiments on, sometimes with the addition of toy problems such as grid worlds. Only a select few papers propose and apply their novel algorithm to methods in different contexts.

Similarly, the types of risk factors we encountered across domains were very different from one another. The concerns of risk in engineering are very different from economics, and so on. This shows that designing a reinforcement learning technique that functions in different contexts is a big challenge in the reinforcement learning literature. We encourage the RL community to tackle this interesting problem.

\subsection{Benchmarks}
There is an emphasis on studies to use toy problems, mainly grid world scenarios, as the means to validate their proposed RL algorithms. We similarly observed a similar trend with certain types of robot control experiments. Tools like Gymnasium (formerly OpenAI Gym) provide benchmarks so that researchers can easily test their creations with other state-of-the-art techniques. These are valuable additions to the RL research community and encourage all future studies to use tools like these, with the objective of making studies easier to compare.

We would also like to suggest that studies should also look beyond these options for their experimental evaluation. As mentioned in the previous point, techniques that have been evaluated on different domains are scarce. The benchmarks featured in Gymnasium and similar consist of grid world, Atari game, and robot and vehicle control problems. While useful for providing results, these are rather simple environments when compared to many real-life problems: safe self-driving cars, operation of real robotic equipment, recommendation systems, and systems that monitor health and administer treatment, among many other potentially useful applications of RL. We believe reinforcement learning studies should start seeking problems with higher complexity as the field has started maturing over time. Such experiments can show limitations to existing techniques and new challenges to overcome; that will eventually result in better methods. Even better, these problems can also eventually become benchmarks available in open-source repositories.

\subsection{Potential Research Areas}
%In the following paragraphs, we highlight a selection of open problems or areas with a scarcity of research, based on the results of our mapping.

Multi-agent reinforcement learning is the main topic of some of the selected studies. This includes both cooperating and competing agents. RL techniques that model the behavior and risks associated with other agents seemed significantly different from other methods. Agents discover rules imposed by the environment. Stochasticity of the environment and the range of exploration affects agents.  The agent is also able to affect the environment itself causing a reaction to the agent's actions. %Some studies have focused on environmental risks, but remain in the minority.

With respect to domains, we expected the economics and medicine domains to have more studies on risk. These domains however do not show up in our mapping. The few papers that do show up relate to risks related to the variance of outcomes. Given the importance of traditional (in the sense of human decision making) risk modeling in these domains, it is important to more systematically characterize the risks of RL systems in these domains.

%economics, and medicine had almost no studies. These problems focus on different types of risk, trending towards the variance of the outcome of an action.

In terms of technology and methodology, we did not encounter any studies that use worst-case criterion, as described in the taxonomy by~\cite{garcia2015comprehensive}. Understandably, these types of techniques can be restrictive in the more common applications. There is the potential use for worst-case criterion techniques where risk has a very heavy severity, like in scenarios where human life is at stake.

We observed a lack of seeded-knowledge exploration approaches in the surveyed papers. Providing initial knowledge to an agent is very useful for real-life applications where the risk holds heavy consequences. We encourage researchers to keep visiting these approaches.

\section{Conclusion}
\label{sec:conc}

This paper presented a systematic literature mapping of risk in the safe RL literature. It proposes a general definition of risk and lists the general attributes of risk across domains that affords an intuitive mapping that can be visually communicated. Moreover, we contributed to the RL literature by showing the trends of recent research in terms of the application domains and type of technique under the risk taxonomy by~\cite{garcia2015comprehensive}.

Our aim is to encourage the research community to adopt an explicit and detailed account of risk in RL approaches. As RL research transfers to more practical applications, it is necessary to articulate more clearly how risk is modeled in RL approaches. Risk-aware techniques can be potentially used in more application domains with only small modifications, thus serving a larger crowd. We hope this mapping is useful for researchers or practitioners considering the implementation of RL in their application domains.

\backmatter

% \bmhead{Supplementary information}

% If your article has accompanying supplementary file/s please state so here. 

% Authors reporting data from electrophoretic gels and blots should supply the full unprocessed scans for key as part of their Supplementary information. This may be requested by the editorial team/s if it is missing.

% Please refer to Journal-level guidance for any specific requirements.

\bmhead{Acknowledgments}

The first author thanks the \textit{University of Costa} Rica (\textit{Universidad de Costa Rica}) for their contributions and support of his Ph.D. program. Martin was partially supported through a grant from the Acquisition Innovation Research Center. Krishnan was partially supported through a grant from the Air Force Office of Research. Gagné was partially supported through a scholarship with the Operations Research department and Northrop Grumman. The views expressed in this paper are those of the authors and do not reflect the official policy or position of the sponsoring agencies.

% \section*{Declarations}

% Some journals require declarations to be submitted in a standardised format. Please check the Instructions for Authors of the journal to which you are submitting to see if you need to complete this section. If yes, your manuscript must contain the following sections under the heading `Declarations':

% \begin{itemize}
% \item Conflict of interest/Competing interests (check journal-specific guidelines for which heading to use)
% \item Ethics approval 
% \item Consent to participate
% \item Consent for publication
% \item Availability of data and materials
% \item Authors' contributions
% \end{itemize}

% \noindent
% If any of the sections are not relevant to your manuscript, please include the heading and write `Not applicable' for that section. 

% \bigskip

% Springer journals and proceedings: \url{https://www.springer.com/gp/editorial-policies}

% Nature Portfolio journals: \url{https://www.nature.com/nature-research/editorial-policies}

% \textit{Scientific Reports}: \url{https://www.nature.com/srep/journal-policies/editorial-policies}

% BMC journals: \url{https://www.biomedcentral.com/getpublished/editorial-policies}
% \end{flushleft}

\begin{appendices}

% \section{Section title of first appendix}\label{secA1}

% An appendix contains supplementary information that is not an essential part of the text itself but which may be helpful in providing a more comprehensive understanding of the research problem or it is information that is too cumbersome to be included in the body of the paper.




\end{appendices}


\bibliography{sn-article}% common bib file

\end{document}